\documentclass{article}

\usepackage{microtype}
\usepackage{graphicx}
\usepackage{subcaption}
\usepackage{booktabs} 
\usepackage{booktabs}
\usepackage{siunitx}
\usepackage{xcolor,colortbl}
\usepackage{booktabs}
\usepackage[table]{xcolor}
\usepackage{tabularx}
\usepackage{makecell}
\usepackage{marvosym}
\usepackage{array} 
\usepackage{booktabs}

\usepackage{multirow}
\usepackage{hyperref}

\definecolor{oursblue}{RGB}{225,240,255}

\usepackage[preprint]{icml2026}

\usepackage{amsmath}
\usepackage{amssymb}
\usepackage{mathtools}
\usepackage{amsthm}
\usepackage{colortbl}
\usepackage{cleveref}
\usepackage{graphicx}
\usepackage{caption}
\usepackage{subcaption}
\usepackage{booktabs}

\definecolor{upColor}{RGB}{17,138,21}
\definecolor{downColor}{RGB}{174,36,67}

\theoremstyle{plain}

\theoremstyle{definition}

\theoremstyle{remark}

\usepackage[textsize=tiny]{todonotes}

\icmltitlerunning{PocketDP3: Efficient Pocket-Scale 3D Visuomotor Policy}

\begin{document}

\twocolumn[
  \icmltitle{
    \mbox{}\protect\makebox[0pt][r]{
        \raisebox{-7pt}{
            \includegraphics[height=0.8cm]{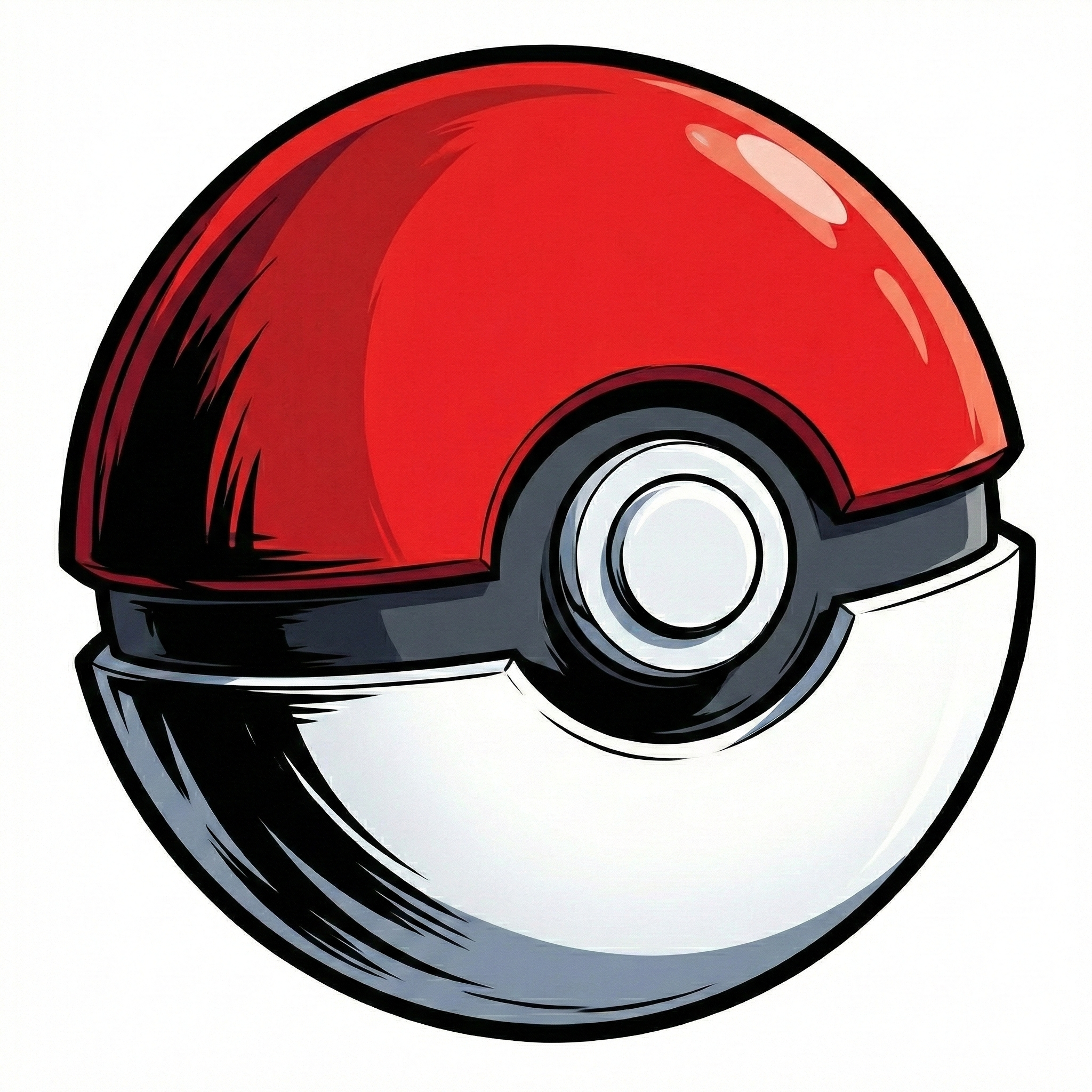}
        }
        \hspace{-0.2cm}
    }PocketDP3: Efficient Pocket-Scale 3D Visuomotor Policy}



  \icmlsetsymbol{equal}{*}

  \begin{icmlauthorlist}
    \icmlauthor{Jinhao Zhang}{equal}
    \icmlauthor{Zhexuan Zhou}{equal}
    \icmlauthor{Huizhe Li}{}
    \icmlauthor{Yichen Lai}{}
    \icmlauthor{Wenlong Xia}{}
    \icmlauthor{Haoming Song}{}
    \icmlauthor{Youmin Gong}{}
    \icmlauthor{Jie Mei\Letter}{}
  \end{icmlauthorlist}


  \icmlcorrespondingauthor{Jie Mei}{jmei@hit.edu.cn}
  

  \icmlkeywords{Machine Learning, ICML}

  \vskip 0.3in
]



\printAffiliationsAndNotice{\icmlEqualContribution}  

\begin{abstract}
    Recently, 3D vision-based diffusion policies have shown strong capability in learning complex robotic manipulation skills. However, a common architectural mismatch exists in these models: a tiny yet efficient point-cloud encoder is often paired with a massive decoder. Given a compact scene representation, we argue that this may lead to substantial parameter waste in the decoder. Motivated by this observation, we propose PocketDP3, a pocket-scale 3D diffusion policy that replaces the heavy conditional U-Net decoder used in prior methods with a lightweight Diffusion Mixer (DiM) built on MLP-Mixer blocks. This architecture enables efficient fusion across temporal and channel dimensions, significantly reducing model size. Notably, without any additional consistency distillation techniques, our method supports two-step inference without sacrificing performance, improving practicality for real-time deployment. Across three simulation benchmarks—RoboTwin2.0, Adroit, and MetaWorld—PocketDP3 achieves state-of-the-art performance with fewer than 1$\%$ of the parameters of prior methods, while also accelerating inference. Real-world experiments further demonstrate the practicality and transferability of our method in real-world settings. Code is available at: \url{https://github.com/jhz1192/PocketDP3.git}.
\end{abstract}

\section{Introduction}
Robotic manipulation has made rapid progress in recent years, largely driven by learning-based approaches that can acquire skills directly from data. Among them, imitation learning (also known as learning from demonstration) has become a particularly practical paradigm: by learning from expert demonstrations, robots can bypass brittle hand-engineered pipelines and directly map observations to actions for contact-rich, long-horizon tasks~\cite{argall2009survey,osa2018algorithmic}. 
However, standard behavior cloning (BC) often reduces policy learning to supervised regression, which can struggle with the inherently \emph{multimodal} nature of human demonstrations (e.g., multiple valid grasps and manipulation strategies), leading to mode-averaging behaviors and compounding errors under distribution shift~\cite{ross2011reduction}.

\begin{figure}[!t]
    \centering
    \includegraphics[width=1.0\linewidth]{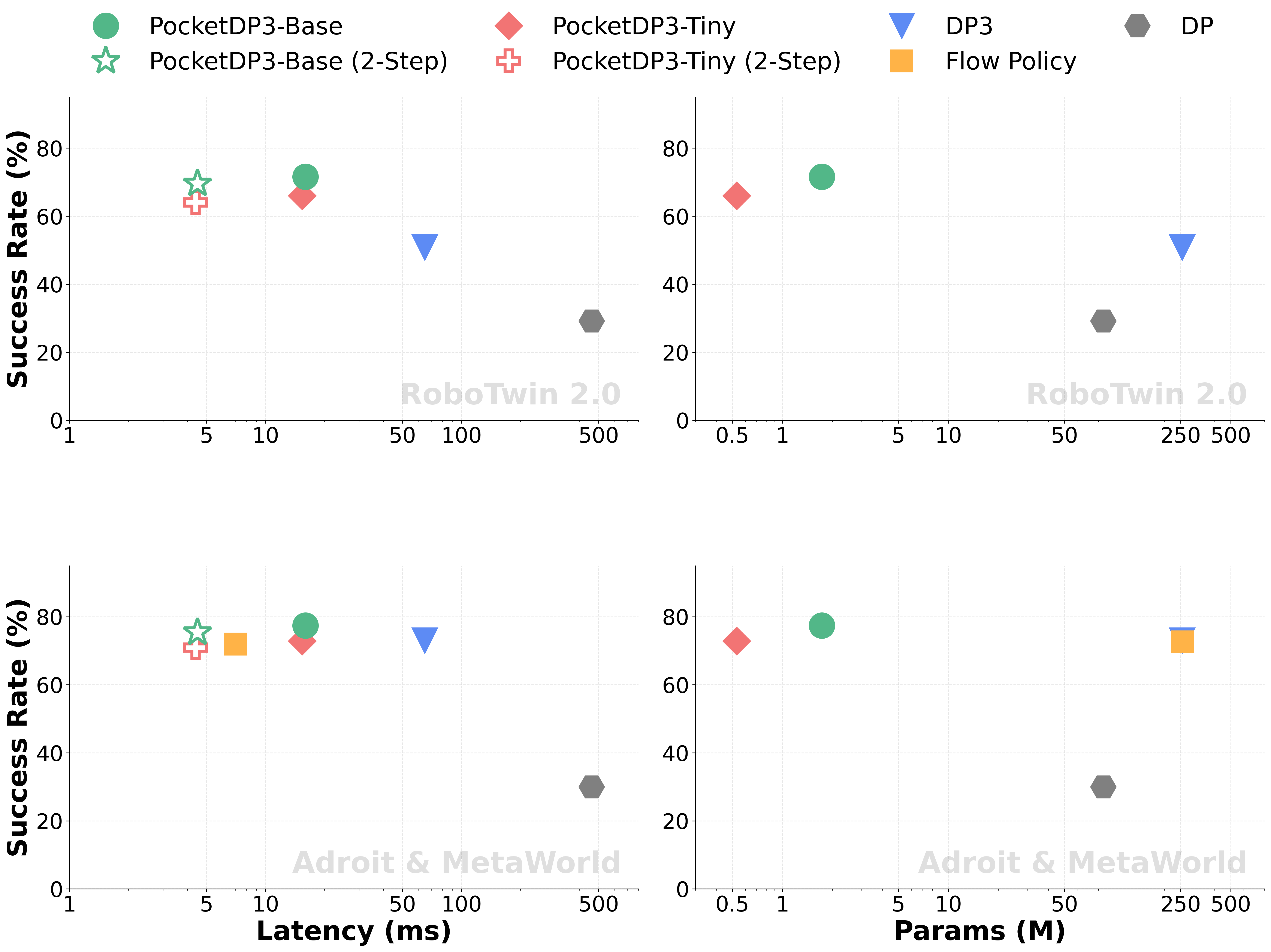}
    \caption{Comparison of PocketDP3 with the state-of-the-art 2D-based method DP\cite{chi2025diffusion} and 3D-based methods DP3\cite{ze20243d}, FlowPolicy\cite{zhang2025flowpolicy} in terms of inference latency/model size and average success rate on RoboTwin2.0, Adroit, and MetaWorld. }
    \label{fig:comp}
    \vspace{-4.0mm}
\end{figure}

Diffusion models offer an appealing alternative by framing generation as a \emph{conditional denoising} process~\cite{sohl2015deep,ho2020denoising}. 
Instead of directly regressing a single action, diffusion-based policies learn to model the full distribution over action trajectories conditioned on observations, enabling expressive multimodal action generation and improved stability in high-dimensional control~\cite{chi2025diffusion}. 
While early diffusion policies were predominantly image-conditioned, recent work has increasingly moved toward 3D representations—most notably point clouds—to improve robustness under viewpoint and lighting changes, and to leverage explicit geometric structure~\cite{ze20243d}.

Despite this progress, we observe a common \emph{architectural mismatch} in 3D diffusion visuomotor policies: a lightweight point-cloud encoder is frequently paired with an extremely large decoder backbone (often a conditional U-Net)~\cite{ronneberger2015u}. 
This design is reasonable for high-resolution image synthesis, where the decoder must reconstruct fine-grained pixel-space structure. In 3D diffusion policies, however, the encoder can already produce a compact yet information-rich scene representation, and the prediction target is a low-frequency trajectory signal(we adopt $x_0$-prediction as in DP3), suggesting that an oversized decoder may be parameter-inefficient. Under these conditions, the strong spatial inductive bias of U-Nets may provide diminishing returns, especially when the conditioning is delivered primarily as a compressed semantic representation. Consequently, decoder-heavy architectures can increase memory footprint and latency—hindering deployment on resource-constrained robots—while offering limited gains in control performance.

\begin{figure*}[htbp]
  \centering
  \includegraphics[width=0.9\textwidth]{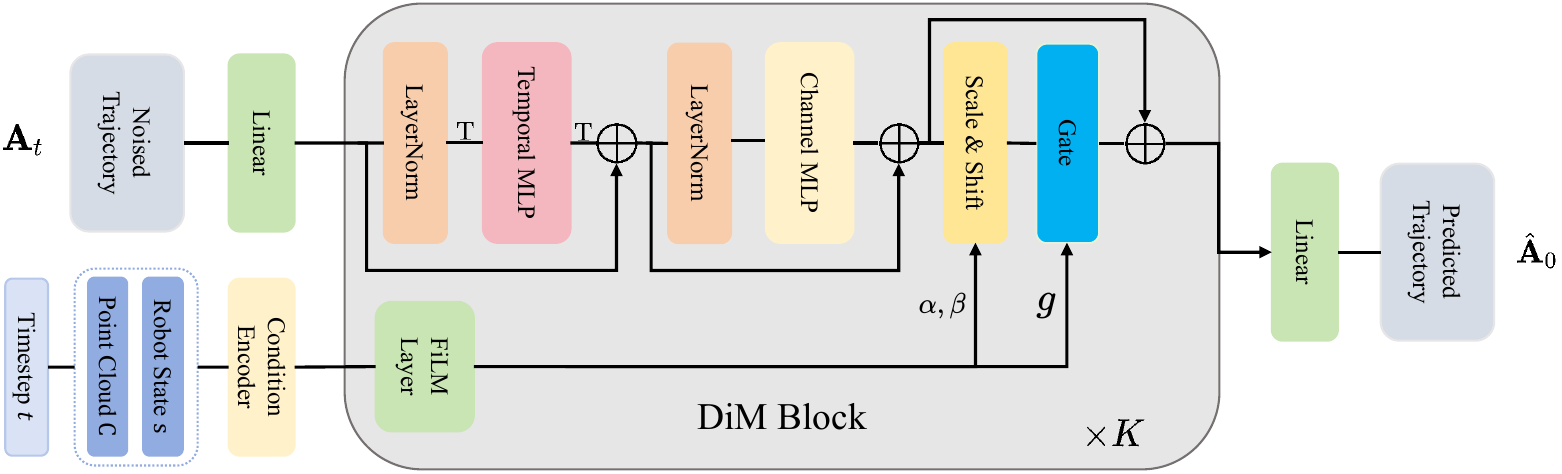}
  \caption{\textbf{Overall architecture of PocketDP3.} 
  In the figure, $\rm T$ denotes transpose.Our PocketDP3 adopts the efficient point-cloud encoder from DP3~\cite{ze20243d} and stacks $K$ DiM blocks as the decoder. Each DiM block is built upon an MLP-Mixer style~\cite{tolstikhin2021mlp} architecture, enabling efficient information fusion with a small parameter budget, thereby improving decision-making performance.}
  \label{fig:dim}
  \vspace{-4.0mm}
\end{figure*}

Motivated by this observation, we ask a simple question: \emph{What is the minimal viable decoder architecture for 3D diffusion-based visuomotor control?} 
We propose \textbf{PocketDP3}, a pocket-scale 3D diffusion policy that substantially reduces decoder size while retaining and often improving performance. Departing from the conventional reliance on U-Net architectures in existing 3D diffusion policies, we explore the efficacy of mixer-based architectures for the first time in this domain. As illustrated in Fig.~\ref{fig:dim}, PocketDP3 retains the lightweight observation encoder of DP3~\cite{ze20243d} while replacing the heavy conditional U-Net decoder with a compact \textbf{Diffusion Mixer (DiM)}.
DiM performs efficient fusion by alternating mixing over temporal and channel dimensions, with lightweight conditioning injection, enabling effective trajectory refinement without a large convolutional decoder. 
Overall, the MLP-Mixer decoder better matches the compressed, global encoder latents and removes spatial redundancy that can hinder existing 3D diffusion policies.


Beyond parameter efficiency, diffusion-based policies face another major challenge: \emph{slow inference}. 
Standard diffusion sampling typically requires many iterative denoising steps, which can be prohibitive for real-time control.
A growing line of work accelerates sampling through distillation and consistency-style training, including progressive distillation and consistency models~\cite{salimans2022progressive,song2023consistency}, as well as robotics-focused variants such as One-Step Diffusion Policy (OneDP)~\cite{wang2025onestep}, Consistency Policy~\cite{prasad2024consistency}, and flow/rectified-flow-inspired policies such as FlowPolicy~\cite{zhang2025flowpolicy}.
In contrast, with DDIM-style sampling~\cite{song2020denoising}, PocketDP3 requires only 2 function evaluations (denoising steps) at inference time, without consistency training or distillation, substantially simplifying training and enabling low-latency deployment while maintaining strong performance.

Our main contributions are summarized as follows:
\begin{itemize}
    \item \textbf{Architectural insight.} We identify a widespread architectural imbalance in 3D diffusion visuomotor policies: compact 3D encoders are often paired with large U-Net decoders, leading to inefficient use of parameters and computation.
    \item \textbf{PocketDP3.} We propose a pocket-scale 3D diffusion policy that replaces the heavy conditional U-Net decoder with a lightweight \textbf{Diffusion Mixer (DiM)} for efficient temporal/channel fusion.
    \item \textbf{Fast sampling without distillation.} We demonstrate two-step inference \emph{without} additional distillation/consistency training, improving practicality for real-time control.
    \item \textbf{Strong empirical results.} PocketDP3 matches or surpasses prior state-of-the-art performance on simulation benchmarks and real-world tasks while substantially reducing parameters and accelerating inference.
\end{itemize}

\section{Related Work}
\textbf{Diffusion Models for Visuomotor Control.} Traditional Behavior Cloning (BC) often struggles with multimodal action distributions, leading to mode-averaging artifacts that result in suboptimal or unsafe behaviors in manipulation tasks. Diffusion Policy \cite{chi2025diffusion} addresses this fundamental limitation by formulating policy learning as a conditional denoising process over the action space. By learning the gradient of the action distribution, it effectively captures complex, multimodal human behaviors and provides superior stability in high-dimensional manipulation tasks compared to explicit regression policies. This probabilistic formulation has demonstrated remarkable expressiveness in capturing the inherent stochasticity of human demonstrations, outperforming prior generative approaches such as Implicit Behavior Cloning \cite{florence2022implicit} and Conditional VAEs \cite{sohn2015learning} across diverse benchmarks. Furthermore, the iterative refinement mechanism inherent to diffusion models enables the generation of temporally coherent action sequences, which is critical for contact-rich manipulation scenarios.

\textbf{From 2D to 3D Representations.} While early diffusion policies relied predominantly on 2D RGB images as visual observations, such representations lack robustness against lighting variations, viewpoint changes, and domain shifts between simulation and real-world environments. Recent works leverage 3D point clouds processed by geometric encoders like PointNet++ \cite{qi2017pointnet++} to extract view-invariant spatial features that better capture object geometry and spatial relationships. 3D Diffusion Policy (DP3) \cite{ze20243d} integrates these efficient 3D representations with diffusion models, achieving state-of-the-art data efficiency and generalization capabilities in few-shot imitation learning settings. Complementary approaches such as Act3D \cite{gervet2023act3d} and PerAct \cite{shridhar2023perceiver} have further demonstrated the benefits of 3D-aware representations for language-conditioned manipulation. However, existing architectures typically pair lightweight point encoders with computationally heavy U-Net decoders for the diffusion backbone, creating a significant parameter redundancy and inference efficiency bottleneck that limits real-time deployment on resource-constrained robotic platforms.

\textbf{Inference Acceleration.} To overcome the prohibitively high latency of iterative denoising during deployment, recent research has focused on principled approaches to sampling acceleration without sacrificing action quality. One-Step Diffusion Policy (OneDP) \cite{wang2025onestep} employs knowledge distillation techniques to compress the multi-step diffusion process into a single forward pass, achieving real-time control frequencies of 62Hz suitable for reactive manipulation. Consistency Policy \cite{prasad2024consistency} takes an alternative approach by enforcing self-consistency constraints along probability flow ordinary differential equation (ODE) trajectories, enabling high-quality action generation in just 1-2 denoising steps while preserving the multimodal expressiveness of the original diffusion formulation. Drawing inspiration from recent advances in generative modeling, FlowPolicy \cite{zhang2025flowpolicy} utilizes Rectified Flow Matching \cite{liu2023flow} to learn straight-line ODE paths between noise and data distributions, effectively avoiding the curvature inherent to standard diffusion processes and enabling faster, more numerically stable inference. These acceleration techniques collectively represent a promising direction toward bridging the gap between the representational power of diffusion-based policies and the stringent latency requirements of closed-loop robotic control.

\section{Methodology}
\subsection{Preliminaries}
\textbf{Diffusion Models.} 
Diffusion models construct a generative process by pairing (i) a fixed \emph{forward} corruption that gradually destroys structure in data and (ii) a learned \emph{reverse} process that progressively restores it. In DDPMs~\cite{ho2020denoising}, the forward process is defined as a Markov chain that adds Gaussian noise at each step:
\begin{equation}
q(x_t\mid x_{t-1})=\mathcal{N}\!\left(x_t;\sqrt{1-\beta_t}\,x_{t-1},\beta_t I\right),
\end{equation}
where $\{\beta_t\}_{t=1}^T$ is a pre-specified noise schedule. As $t$ increases, the distribution of $x_t$ becomes increasingly noisy and approaches an isotropic standard Gaussian. A convenient property of this construction is that the marginal at any timestep has a closed form:
\begin{equation}
q(x_t\mid x_0)=\mathcal{N}\!\left(x_t;\sqrt{\bar{\alpha}_t}\,x_0,(1-\bar{\alpha}_t)I\right),
\end{equation}
with $\alpha_t=1-\beta_t$ and $\bar{\alpha}_t=\prod_{s=1}^t \alpha_s$.

Generation is performed by running a learned reverse-time chain initialized from the terminal prior $x_T\sim\mathcal{N}(0,I)$. DDPMs model each reverse transition with a Gaussian:
\begin{equation}
p_\theta(x_{t-1}\mid x_t)=\mathcal{N}\!\left(x_{t-1};\mu_\theta(x_t,t),\Sigma_\theta(x_t,t)\right).
\end{equation}
In practice, instead of parameterizing $\mu_\theta$ directly, the model is often trained to predict the noise that produced $x_t$ from $x_0$ using a neural network $\epsilon_\theta(x_t,t)$. This yields the widely used denoising objective
\begin{equation}
\mathcal{L}=\mathbb{E}_{t,x_0,\epsilon}\left[\left\|\epsilon-\epsilon_{\theta}\!\left(\sqrt{\bar{\alpha}_t}x_0+\sqrt{1-\bar{\alpha}_t}\,\epsilon,t\right)\right\|_2^2\right],
\end{equation}
which is equivalent to optimizing a variational lower bound of the data log-likelihood.

While DDPM sampling follows the learned stochastic reverse chain, we adopt DDIM-style~\cite{song2020denoising} sampling in our framework, which enables fewer denoising steps while maintaining high generation quality, making it well-suited for integration into our policy architecture.

\textbf{Problem Formulation.}
We consider a dataset of expert demonstrations consisting of observation--action pairs. At each frame, the observation comprises a single-view, robot-centric point cloud ${\bf P}$ and the robot proprioceptive state ${\bf s}$, and the associated expert action serves as the supervision signal. Our goal is to learn an end-to-end visuomotor policy $\pi_\theta : \mathcal{O} \to \mathcal{A}$ parameterized by $\theta$, where $\mathcal{O}$ and $\mathcal{A}$ denote the observation space and action space, respectively. Such a policy should not only learn complex manipulation skills, but also generalize across diverse scenes.

\subsection{Overview}
Although diffusion-based methods that leverage 3D information can already generate high-quality robot actions~\cite{ze20243d, zhang2025flowpolicy}, we argue that current approaches exhibit a fundamental mismatch: a tiny yet efficient point-cloud encoder is paired with a massive U-Net decoder(e.g., DP3 uses a $\sim$0.05M-parameter conditional encoder but a $\sim$250M-parameter U-Net decoder). Since the point-cloud encoder provides a dense and information-rich scene representation, an oversized decoder may introduce substantial parameter redundancy.

In this paper, we argue that while larger decoders can improve denoising capacity~\cite{dhariwal2021diffusion}, given a dense and efficient scene representation, a much smaller model can match or even outperform previous “large-decoder” policies, provided that the fusion mechanism is properly designed. Accordingly, our goal is to identify the "minimal viable" decoder for 3D visuomotor policies, thereby improving their practicality for real-time applications.

As shown in Fig. \ref{fig:dim}, we propose to replace the U-Net decoder backbone with a simple and efficient fusion architecture, MLP-Mixer \cite{tolstikhin2021mlp}. We retain DP3’s observation encoder, but replace its conditional U-Net~\cite{ronneberger2015u} decoder with our proposed Diffusion Mixer (DiM). This design matches, and in some cases surpasses, the original model’s performance while using less than 1\% of the parameters.

\textbf{Observation Encoder.}
We adopt the same lightweight observation encoder as DP3~\cite{ze20243d}. Specifically, we first downsample the input point cloud via farthest point sampling (FPS), which preserves spatial coverage while reducing redundancy. The downsampled points are then processed by a simple MLP followed by a max-pooling operation to obtain a compact 3D representation $\boldsymbol{z}$, which serves as the visual conditioning signal. We linearly project the robot state $s$ and add it to $\boldsymbol{z}$ to form the denoising context:
\begin{equation}
C = \boldsymbol{z} + \mathrm{Linear}(s).
\end{equation}
\textbf{DiM Decoder.}
The DiM decoder takes the noisy trajectory ${\bf A}_t$ as input, and is conditioned on the scene representation $z$ and the diffusion timestep $t$. Following the $x_0$-prediction parameterization~\cite{ramesh2022hierarchical}, it directly outputs the predicted clean trajectory $\hat{\bf A}_0$. Specifically, the noisy trajectory ${\bf A}_t$ is first projected by a linear layer into a sequence of tokens. These tokens then interact and fuse with the scene representation $z$ and the diffusion timestep $t$ through $K$ DiM blocks. Finally, another linear layer maps the fused tokens to the final prediction $\hat{\bf A}_0$. The detailed design of the DiM block is provided in Sec. \ref{sec:dim_block}.

\textbf{Training and Sampling.}
During training, we first sample a clean trajectory ${\bf A}_0$ and its corresponding observation $(P,s)$ from the dataset. We then apply one step noising to ${\bf A}_0$ to obtain the noisy data ${\bf A}_t$:
\begin{equation}
    {\bf A}_t = \sqrt{\bar{\alpha}_t} {\bf A}_0 + \sqrt{1-\bar{\alpha}_t} \boldsymbol{\epsilon}, \quad \boldsymbol{\epsilon} \sim \mathcal{N}({\bf 0}, {\bf I})
\end{equation}
where $\bar{\alpha}_t$ are noise schedule that performs one step noise adding~\cite{ho2020denoising}. We then train a neural network $\boldsymbol{\mu}_{\theta}$ to predict the original data $\hat{\bf A}_0=\boldsymbol{\mu}_{\theta}(\hat{\bf A}_t,t,P,s)$:
\begin{equation}
    \mathcal{L}=\mathbb{E}_{({\bf A}_0,P,s),t,\epsilon}\left[ \|\hat{\bf A}_0-{\bf A}_0\|^2  \right]
\end{equation}
At inference time, we use DDIM~\cite{song2020denoising} as the noise scheduler to accelerate sampling. Notably, we find that without any consistency distillation techniques~\cite{prasad2024consistency, wang2024one, zhang2025flowpolicy}, we can achieve two-step inference without sacrificing performance.

\subsection{Design of Diffusion Mixer Block} \label{sec:dim_block}
Most prior work adopts a conditional U-Net as the decoder~\cite{chi2025diffusion, ze20243d, zhang2025flowpolicy}. This CNN-based architecture typically relies on wider channel widths to improve denoising performance~\cite{zagoruyko2016wide}. While it has demonstrated strong capability in robotic manipulation, increasing the channel width often results in a massive parameter count, which is mismatched with the lightweight but strong point-cloud encoder. To improve information fusion without increasing model size, we employ MLP-Mixer~\cite{tolstikhin2021mlp} as a simple yet effective backbone. With only inexpensive transpose operations and MLP-based projections, MLP-Mixer enables efficient cross-channel interaction and improves training stability.

As shown in Fig. \ref{fig:dim}, given the token sequence $H$ obtained by projecting the noisy trajectory, we first transpose it and feed it into a temporal MLP to aggregate information across the time dimension. We then transpose it back and add a residual connection. Next, we apply the same procedure along the channel dimension, resulting in a representation that fuses both temporal and channel-wise information:
\begin{equation}
    H = H+{\rm MLP}(H^{\rm T})^{\rm T};\quad H = H + {\rm MLP}(H)
\end{equation}
Finally, we inject the conditioning information into the model using a FiLM layer combined with a gated residual connection:
\begin{gather}
    \alpha, \beta, g = {\rm MLP}(H, C) \\
    \tilde{H} = H \odot \alpha + \beta \\
    H = \tilde{H} \odot g  + H
\end{gather}
After stacking $K$ such blocks, we obtain the final tokens in which the conditioning information is fully mixed, which are then used to predict the denoised actions:
\begin{equation}
    \hat{\bf A}_0 = {\rm Linear}(H)
\end{equation}

\begin{table}[t]
\centering 
\caption{{\bf The detailed configuration of PocketDP3.}}
\label{tab:mc}
\scriptsize
\begin{tabular}{lccc} 
\toprule
\textbf{Model} &
\textbf{Layers} $K$ &
\textbf{Hidden size} $d$ & 
\textbf{Num. Params.}(M)  \\
\midrule
PocketDP3-tiny  & 4 & 64 & 0.53 \\
PocketDP3-base & 4 & 128 & 1.73 \\
\bottomrule
\end{tabular}
\vspace{-0.3cm}
\end{table}

\textbf{Model size.}
In our experiments, we evaluate two model configurations, Tiny and Base, with detailed hyperparameters reported in Tab. \ref{tab:mc}. Base version has more parameters (still only 0.67$\%$ of DP3), and achieves better performance. Tiny version, with just 0.53M parameters (0.21$\%$ of DP3), attains performance competitive to Base while substantially outperforming DP3 (see the Sec.\ref{sec:exp} for detailed results).

\begin{table*}[t]
\centering
\caption{Evaluation on the Robotwin2.0 benchmark. Each task is tested across 100 randomly generated scenes using 100 different seeds.}
\label{tab:robotwin2}

\small
\setlength{\tabcolsep}{4.5pt}
\renewcommand{\arraystretch}{1.15}

\begin{tabularx}{\textwidth}{@{}l *{4}{>{\centering\arraybackslash}X}@{}}
\toprule
\textbf{Method} & \mbox{Beat Block Hammer} & \mbox{Click Alarmclock} & \mbox{Click Bell} & \mbox{Hanging Mug} \\
\midrule
DP & 42.0 & 61.0 & 54.0 & 8.0 \\
$\pi_0$ & 43.0 & 63.0 & 44.0 & 11.0 \\
DP3 & 72.0 & 77.0 & \underline{90.0} & 17.0 \\
\rowcolor{oursblue}
PocketDP3-tiny \textbf{(Ours)} & \textbf{94.0}\,{\scriptsize(+22.0)} & \underline{94.0}\,{\scriptsize(+17.0)} & \textbf{100.0}\,{\scriptsize(+10.0)} & \underline{29.0}\,{\scriptsize(+12.0)} \\
\rowcolor{oursblue}
PocketDP3-base \textbf{(Ours)} & \underline{92.0}\,{\scriptsize(+20.0)} & \textbf{98.0}\,{\scriptsize(+21.0)} & \textbf{100.0}\,{\scriptsize(+10.0)} & \textbf{39.0}\,{\scriptsize(+22.0)} \\

\addlinespace[3pt]
\midrule
\textbf{Method} & \mbox{Move Can Pot} & \mbox{Move Pillbottle Pad} & \mbox{Pick Diverse Bottles} & \mbox{Pick Dual Bottles} \\
\midrule
DP & 39.0 & 1.0 & 6.0 & 24.0 \\
$\pi_0$ & 58.0 & 21.0 & 27.0 & 57.0 \\
DP3 & 70.0 & 41.0 & 52.0 & 60.0 \\
\rowcolor{oursblue}
PocketDP3-tiny \textbf{(Ours)} & \underline{81.0}\,{\scriptsize(+11.0)} & \underline{45.0}\,{\scriptsize(+4.0)} & \textbf{77.0}\,{\scriptsize(+25.0)} & \underline{83.0}\,{\scriptsize(+23.0)} \\
\rowcolor{oursblue}
PocketDP3-base \textbf{(Ours)} & \textbf{97.0}\,{\scriptsize(+27.0)} & \textbf{53.0}\,{\scriptsize(+12.0)} & \underline{74.0}\,{\scriptsize(+22.0)} & \textbf{93.0}\,{\scriptsize(+33.0)} \\

\addlinespace[3pt]
\midrule
\textbf{Method} & \mbox{Place Bread Basket} & \mbox{Place Bread Skillet} & \mbox{Place Burger Fries} & \mbox{Place Cans Plasticbox} \\
\midrule
DP & 14.0 & 11.0 & 72.0 & 40.0 \\
$\pi_0$ & 17.0 & 23.0 & \underline{80.0} & 34.0 \\
DP3 & 26.0 & 19.0 & 72.0 & 48.0 \\
\rowcolor{oursblue}
PocketDP3-tiny \textbf{(Ours)} & \textbf{48.0}\,{\scriptsize(+22.0)} & \underline{38.0}\,{\scriptsize(+15.0)} & 74.0\,{\scriptsize(-6.0)} & \textbf{98.0}\,{\scriptsize(+50.0)} \\
\rowcolor{oursblue}
PocketDP3-base \textbf{(Ours)} & \underline{41.0}\,{\scriptsize(+15.0)} & \textbf{52.0}\,{\scriptsize(+29.0)} & \textbf{86.0}\,{\scriptsize(+6.0)} & \underline{95.0}\,{\scriptsize(+47.0)} \\

\addlinespace[3pt]
\midrule
\textbf{Method} & \mbox{Place Empty Cup} & \mbox{Place Object Stand} & \mbox{Place Phone Stand} & \mbox{Scan Object} \\
\midrule
DP & 37.0 & 22.0 & 13.0 & 9.0 \\
$\pi_0$ & 37.0 & 36.0 & 35.0 & 18.0 \\
DP3 & 65.0 & 60.0 & 44.0 & \underline{31.0} \\
\rowcolor{oursblue}
PocketDP3-tiny \textbf{(Ours)} & \underline{90.0}\,{\scriptsize(+25.0)} & \underline{61.0}\,{\scriptsize(+1.0)} & \underline{55.0}\,{\scriptsize(+11.0)} & \underline{31.0}\,{\scriptsize(+0.0)} \\
\rowcolor{oursblue}
PocketDP3-base \textbf{(Ours)} & \textbf{94.0}\,{\scriptsize(+29.0)} & \textbf{72.0}\,{\scriptsize(+12.0)} & \textbf{63.0}\,{\scriptsize(+19.0)} & \textbf{43.0}\,{\scriptsize(+12.0)} \\

\addlinespace[3pt]
\midrule
\textbf{Method} & \mbox{Stack Bowls Three} & \mbox{Stamp Seal} & \mbox{Turn Switch} & \mbox{Average} \\
\midrule
DP & 63.0 & 2.0 & 36.0 & 29.2 \\
$\pi_0$ & \underline{66.0} & 3.0 & 27.0 & 36.8 \\
DP3 & 57.0 & 18.0 & 46.0 & 50.8 \\
\rowcolor{oursblue}
PocketDP3-tiny \textbf{(Ours)} & 63.0\,{\scriptsize(-3.0)} & \underline{34.0}\,{\scriptsize(+16.0)} & \textbf{59.0}\,{\scriptsize(+13.0)} & \underline{66.0}\,{\scriptsize(+15.2)} \\
\rowcolor{oursblue}
PocketDP3-base \textbf{(Ours)} & \textbf{72.0}\,{\scriptsize(+6.0)} & \textbf{41.0}\,{\scriptsize(+23.0)} & \underline{56.0}\,{\scriptsize(+10.0)} & \textbf{71.6}\,{\scriptsize(+20.8)} \\

\bottomrule
\end{tabularx}
\end{table*}

\begin{table*}[t]
\centering
\caption{Evaluation on the Adroit and MetaWorld benchmark. Each task is trained and tested across 3 different seeds. Unavailable results are denoted by ``--''.}
\label{tab:am}

\footnotesize
\setlength{\tabcolsep}{4pt}
\renewcommand{\arraystretch}{1.12}

\resizebox{\textwidth}{!}{%
\begin{tabular}{@{}l*{11}{c}@{}}
\toprule
\textbf{Method}
& \textbf{Hammer} & \textbf{Door} & \textbf{Pen}
& \textbf{Assembly} & \textbf{Disassemble} & \textbf{Hand-Insert}
& \textbf{Pick-Place-Wall} & \textbf{Push} & \textbf{Reach-Wall} & \textbf{Stick-Push}
& \textbf{Avg.} \\
\midrule

BCRNN
& 0.0 & 0.0 & 9.0
& 3.0 & 32.0 & --
& -- & -- & -- & --
& 8.8 \\
IBC
& 0.0 & 0.0 & 9.0
& 0.0 & 1.0 & --
& -- & -- & -- & --
& 2.0 \\
DP
& 45.0 & 37.0 & 13.0
& 15.0 & 43.0 & 9.0
& 5.0 & 11.0 & 59.0 & \underline{63.0}
& 30.0 \\

DP3
& \textbf{100.0} & \underline{62.0} & 43.7
& \underline{99.6} & 75.0 & \underline{25.3}
& 82.7 & 71.3 & 70.7 & \textbf{100.0}
& \underline{73.0} \\

Flow Policy
& \textbf{100.0} & \textbf{65.3} & \underline{48.2}
& 92.7 & 60.7 & 18.6
& \underline{88.3} & 79.7 & \underline{72.7} & \textbf{100.0}
& 72.6 \\

\rowcolor{oursblue}
PocketDP3-tiny \textbf{(Ours)}
& \underline{99.7} & 49.7 & 40.7
& 96.3 & \underline{77.7} & 13.3
& 88.0 & \textbf{85.7} & \textbf{78.3} & \textbf{100.0}
& 72.9 \\

\rowcolor{oursblue}
PocketDP3-base \textbf{(Ours)}
& \textbf{100.0} & 54.3 & \textbf{48.7}
& \textbf{100.0} & \textbf{87.7} & \textbf{33.3}
& \textbf{88.7} & \underline{83.0} & \textbf{78.3} & \textbf{100.0}
& \textbf{77.4} \\
\bottomrule
\end{tabular}
}

\vspace{-0.5mm}
\end{table*}

\begin{figure}[ht]
\centering
\includegraphics[width=\linewidth]{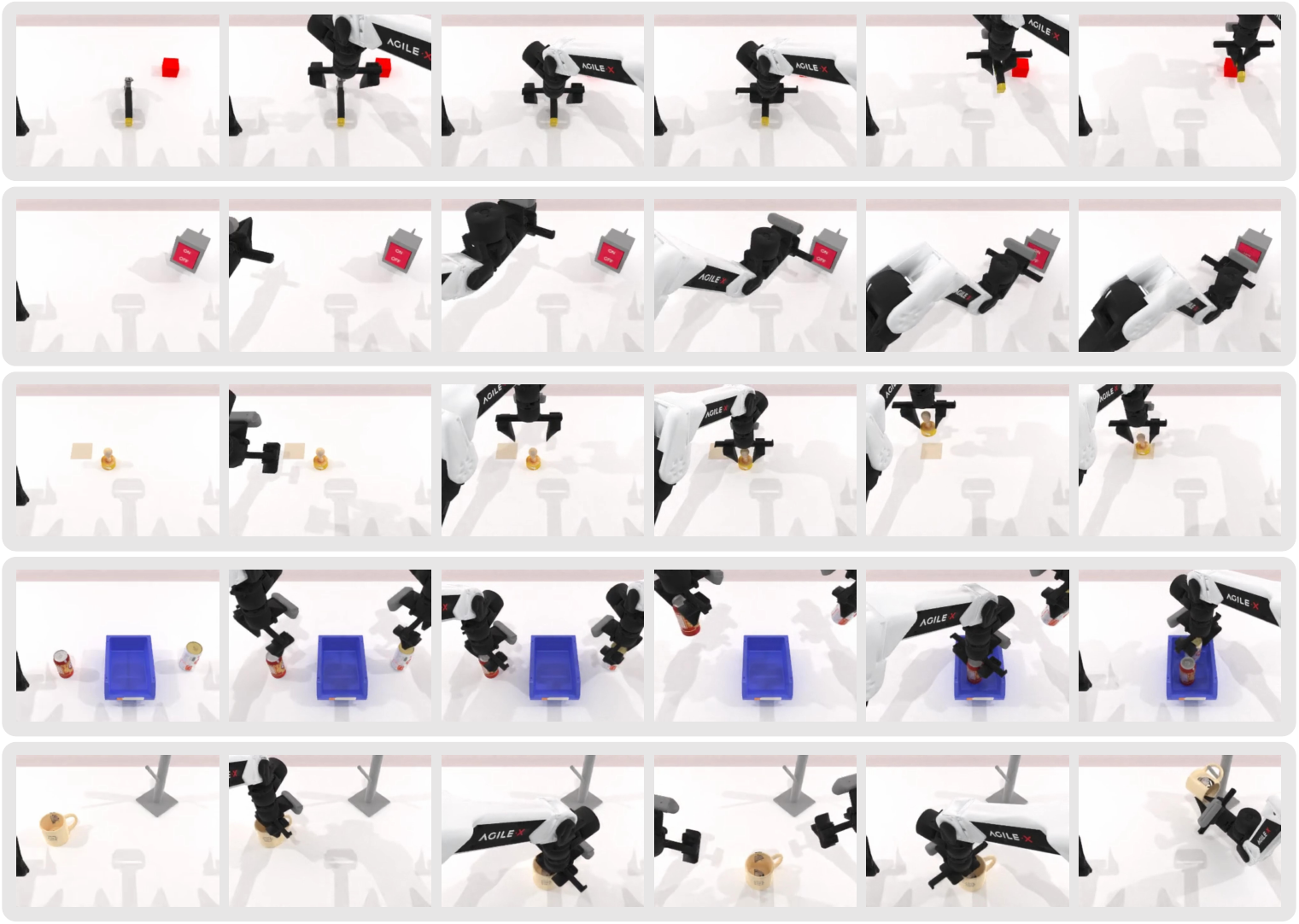}
\caption{\textbf{Visualizations of the simulation experiments.} The simulation results show that our PocketDP3 achieves strong performance across a variety of tasks, demonstrating the efficiency of its architecture design.}
\vspace{-0.3cm}      
\label{fig:sim}
\end{figure}

\section{Experiments} \label{sec:exp}
\subsection{Experimental Setup}
\textbf{Simulation Benchmarks.}
We evaluate our method on three widely used benchmarks: RoboTwin2.0~\cite{chen2025robotwin}, Adroit~\cite{rajeswaran2017learning}, and MetaWorld~\cite{yu2020meta}. To ensure a rigorous comparison and verify that performance gains stem solely from our architectural efficiency, we strictly adhere to the experimental protocols of prior state-of-the-art methods (e.g., DP3~\cite{ze20243d}), utilizing identical datasets and explicitly refraining from any additional data augmentation. RoboTwin2.0 focuses on dual-arm manipulation with diverse assets and scripted data generation. Adroit targets high-dimensional dexterous control for precise, long-horizon skills, while MetaWorld offers tiered single-arm tasks~\cite{seo2023masked}. Collectively, these benchmarks cover both single- and dual-arm settings across varying dimensionalities, ensuring a comprehensive evaluation.

\textbf{Training and Evaluation Details.}
For a fair comparison, we follow the official training and evaluation protocols for all benchmarks. 
\begin{itemize}
    \item For RoboTwin2.0, we train our model using 50 expert demonstrations, following the official guidelines. We evaluate each task on 100 randomly generated scenes and report the success rate over these 100 evaluation episodes.

    \item  For Adroit and MetaWorld, we use 10 expert demonstrations to assess the data efficiency of our method. We run three independent trials with random seeds ${0,1,2}$. During training, we evaluate the policy every 200 epochs; each evaluation consists of 20 rollouts per task, from which we compute the success rate. For each seed, we track the success rate over training and define ${\rm SR}_5$ as the average of the top five success rates; we report the average ${\rm SR}_5$ across the three seeds for all methods.
\end{itemize}

We train our PocketDP3 on a single NVIDIA RTX 5880 Ada, with all key training and inference configurations(except for the number of inference steps) kept identical to the official deployment. We adopt a DDIM~\cite{song2020denoising} noise scheduler with 100 diffusion steps during training and 2 steps at inference, and optimize using AdamW~\cite{loshchilov2017decoupled} with an initial learning rate of $1\times10^{-4}$ and a cosine decay schedule. To stabilize training, both actions and robot states are normalized to $[-1,1]$. All models are trained for 3{,}000 epochs, with a batch size of 256 on RoboTwin2.0 and 128 on Adroit and MetaWorld. Detailed hyperparameter settings are provided in Appendix~\ref{app:impl}.

\textbf{Baselines.}
On RoboTwin2.0, we compare our method with the top-performing approaches on the official leaderboard\footnote{\url{https://robotwin-platform.github.io/leaderboard}}
, namely DP3\cite{ze20243d}, DP\cite{chi2025diffusion}, and the VLA model $\pi_0$\cite{black2024pi_0}.
On Adroit and MetaWorld, we select representative and competitive diffusion-based methods—FlowPolicy\cite{zhang2025flowpolicy}, DP3\cite{ze20243d}, and DP\cite{chi2025diffusion}—and further include commonly used imitation-learning baselines, IBC\cite{florence2022implicit} and BCRNN\cite{mandlekar2021matters}.

\subsection{Comparison with State-of-the-art Methods}
\textbf{RoboTwin 2.0.}
As shown in Tab.~\ref{tab:robotwin2}, both PocketDP3 variants—Tiny and Base—achieve state-of-the-art success rates on RoboTwin2.0. With only \textbf{two inference steps}, PocketDP3-Tiny and PocketDP3-Base outperform the previous SOTA method DP3 by \textbf{15.2}$\%$ and \textbf{20.8}$\%$ in average success rate, respectively, highlighting the effectiveness of our proposed DiM architecture.
Notably, on \textit{Place Cans Plasticbox}, \textit{Pick Dual Bottles}, and \textit{Place Bread Skillet}, PocketDP3-Base improves over DP3 by 47$\%$, 33$\%$, and 33$\%$, respectively, indicating particularly large gains on these tasks. We also observe that while Base performs better on most tasks, Tiny is superior on a few tasks (e.g., \textit{Turn Switch} and \textit{Place Bread Basket}). This supports our hypothesis that overly large models may underutilize capacity and can even amplify optimization noise on certain tasks, resulting in degraded performance. 

\textbf{Adroit and MetaWorld.}
Tab. \ref{tab:am} compares our method with prior SOTA across 10 manipulation tasks on Adroit and MetaWorld. Our PocketDP3-base still achieves the best overall performance with 77.4$\%$ average success, outperforming DP3 (73.0$\%$) and Flow Policy (72.6$\%$). Despite using an extremely small number of parameters and only two inference steps, PocketDP3 maintains---and in some cases surpasses---the performance of prior state-of-the-art methods.

\textbf{Model Size and Inference Latency.}
We measure the model size and inference latency of several diffusion-based policies; the results are reported in Tab. \ref{tab:efficiency}. Thanks to two-step inference and a compact model size, our PocketDP3 substantially reduces inference latency compared to the previous state-of-the-art method, DP3, while maintaining strong performance. Moreover, while FlowPolicy achieves impressive latency via 1-step inference, it retains a massive parameter count ($\sim$255M). PocketDP3 achieves comparable low latency ($\sim$4.5ms) but with \textbf{two orders of magnitude fewer parameters}, significantly lowering the memory footprint for on-device deployment.

\begin{table}[ht]
\centering
\caption{\textbf{Model size and inference latency comparison.} All evaluations are conducted on a single RTX~5880 Ada GPU with the batch size set to 1.}
\label{tab:efficiency}
\footnotesize
\setlength{\tabcolsep}{4pt}
\renewcommand{\arraystretch}{1.1}
\begin{tabularx}{\columnwidth}{@{}Xccc@{}}
\toprule
\textbf{Method} & \textbf{Params (M)$\downarrow$} & \textbf{NFE} & \textbf{Latency (ms)$\downarrow$} \\
\midrule
DP & 85.6 & 100 & 460 \\
DP3 & 255.1 & 10 & 51.4 \\
PocketDP3-base \textbf{(Ours)} & 1.73 & 10 & 16.01 \\
PocketDP3-tiny \textbf{(Ours)} & 0.53 & 10 & 14.93 \\
Flow Policy & 255.8 & \textbf{1} & 7.04 \\
PocketDP3-base \textbf{(Ours)} & 1.73 & 2 & \textbf{4.80} \\
PocketDP3-tiny \textbf{(Ours)} & \textbf{0.53} & 2 & \textbf{4.23} \\
\bottomrule
\end{tabularx}
\vspace{-3.0mm}
\end{table}





\subsection{Ablation Studies}
\textbf{Necessity of the Mixer-based Architecture.}
To validate the design of PocketDP3, we conduct an ablation study that compares different architectures under the same parameter budget. The results are summarized in Tab.~\ref{tab:ablation}. Here, Vanilla-MLP denotes a pure MLP architecture with FiLM-based conditioning, while Vanilla-UNet is obtained by reducing the intermediate channel width in the decoder of the original U-Net. We observe that the pure MLP baseline achieves an average ${\rm SR}_5$ of $0$ across all three tasks. Specifically, the training loss for Vanilla-MLP converged extremely poorly, indicating that a plain MLP is far from sufficient for robotic manipulation. Notably, our DiM block---which augments the MLP with residual connections and a temporal fusion module (incurring almost no additional parameters)---substantially improves performance. 
Furthermore, the U-Net variant with reduced channel width suffers a significant performance drop compared to the original DP3 and is substantially outperformed by our PocketDP3. These results suggest that, under a strict parameter budget, the DiM block offers superior parameter efficiency and information fusion capabilities compared to standard U-Net architectures.

\begin{table}[h]
\centering
\caption{\textbf{Ablation on design choices in PocketDP3.} To ensure a fair comparison, we tune the hidden size for each setting to keep the number of parameters comparable.}
\label{tab:ablation}
\resizebox{\columnwidth}{!}{%
\begin{tabular}{l|c|cc|cc|cc|cc}
\toprule
\textbf{Method} & \textbf{Num.Params.(M)} &
\multicolumn{2}{c|}{\textbf{Door}} &
\multicolumn{2}{c|}{\textbf{Hammer}} &
\multicolumn{2}{c|}{\textbf{Pen}} &
\multicolumn{2}{c}{\textbf{Avg.}} \\
\cmidrule(lr){3-4}\cmidrule(lr){5-6}\cmidrule(lr){7-8}\cmidrule(lr){9-10}
 &  & \textbf{${\rm SR}_5$} & Loss & \textbf{${\rm SR}_5$} & Loss & \textbf{${\rm SR}_5$} & Loss & \textbf{${\rm SR}_5$} & Loss \\
\midrule
PocketDP3-base & \textbf{1.73} &
\textbf{54.3} & 6e-4 &
\textbf{100} & 2e-4 &
\textbf{48.7} & 5e-4 &
\textbf{67.7} & 4.3e-4 \\
Vanilla-MLP    & \textbf{1.73} &
0.0 & 3e-2 &
0.0 & 1e-2 &
0.0 & 3e-2 &
0.0 & 2.3e-2 \\
Vanilla-UNet   & 1.85 &
25.7 & 3e-3 &
93.3 & 3e-4 &
42.0 & 1e-3 &
53.7 & 1.4e-3 \\
\bottomrule
\end{tabular}}
\vspace{-0.2cm}
\end{table}

\textbf{Impact of the Number of Function Evaluations(NFE).}
We further study how the number of inference steps affects the final success rate on three Adroit tasks—\textit{Door}, \textit{Hammer}, and \textit{Pen}. Specifically, we evaluate the final checkpoint after 3000 training epochs. For each NFE setting, we run 5{,}000 evaluation episodes to reduce statistical variance and report the mean score. The results are shown in Tab.~\ref{tab:ablation_nfe}. As can be seen, using only one inference step leads to a noticeable performance drop compared to multi-step settings. However, two-step inference already achieves strong performance, and additional steps (e.g., 5 or 10) do not yield further gains. This suggests that PocketDP3 can meet the requirements of most tasks with just two denoising steps.


\begin{table}[h]
\centering
\caption{\textbf{Ablation on NFE.} We report the average success rate (\%) over 5,000 randomly generated scenes for each task using the final checkpoint (3,000 epochs). \textit{NFE} denotes the Number of Function Evaluations.}
\label{tab:ablation_nfe}
\small 
\setlength{\tabcolsep}{12pt} 
\begin{tabular}{lcccc} 
\toprule
\textbf{NFE} & \textbf{Door} & \textbf{Hammer} & \textbf{Pen} & \textbf{Avg.} \\
\midrule
1  & 34.2 & 91.5 & 33.4 & 53.0 \\
2  & 37.4 & 100.0 & 37.2 & 58.2 \\
5  & 37.7 & 100.0 & 35.5 & 57.7 \\
10 & 36.8 & 100.0 & 37.3 & 58.0 \\
\bottomrule
\end{tabular}
\vspace{-3.0mm}
\end{table}

\subsection{Real-world Experiments}
\textbf{Experiment Settings.}
We validate the effectiveness of our method on an AgileX Piper robot. Real-world visual observations are captured with a single Intel RealSense D455 camera mounted globally. The model is deployed on an NVIDIA RTX 4060 GPU for on-board action inference. We collect expert demonstrations via teleoperation using a leader-follower setup, yielding 50 trajectories for training. The action space is defined as 6-DoF joint positions.

\begin{figure}[!t]
    \centering
    \includegraphics[width=0.8\linewidth]{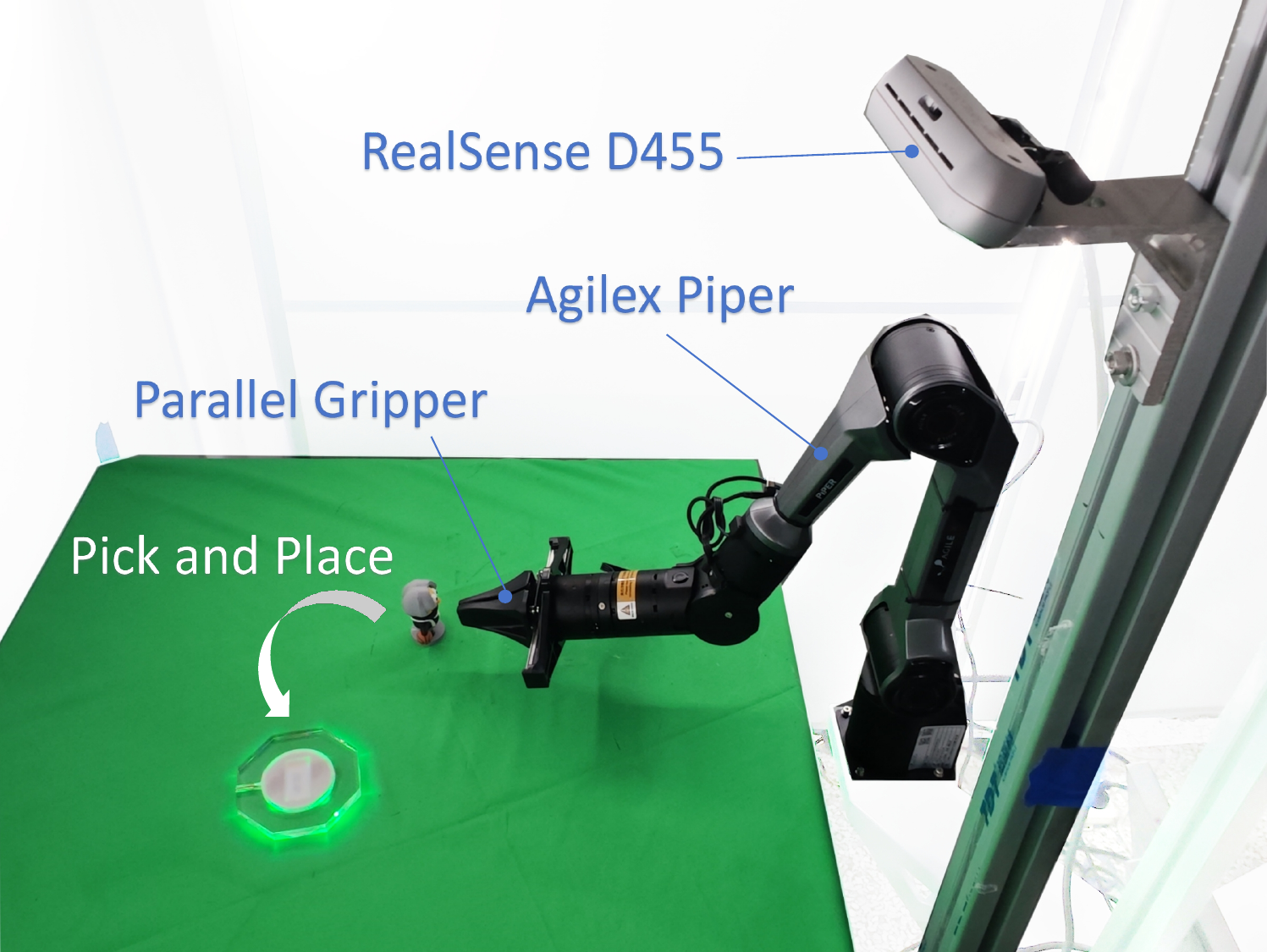}
    \caption{\textbf{Real-World Experiment Setup.} }
    \label{fig:task}
    \vspace{-3.0mm}
\end{figure}


\begin{figure}[ht]
  \centering
  \includegraphics[width=\linewidth]{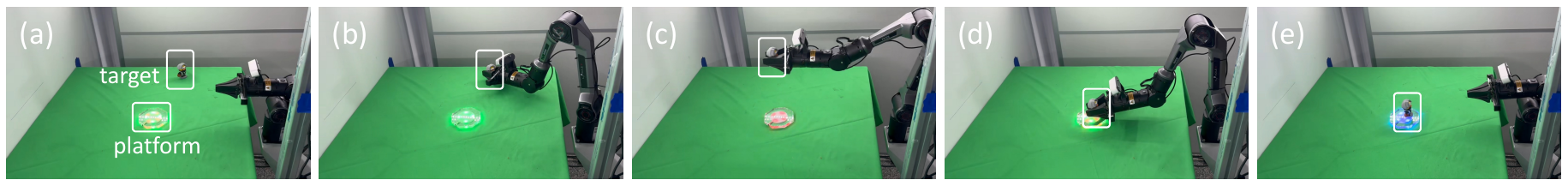} \\
  \vspace{1pt} 
  
  \includegraphics[width=\linewidth]{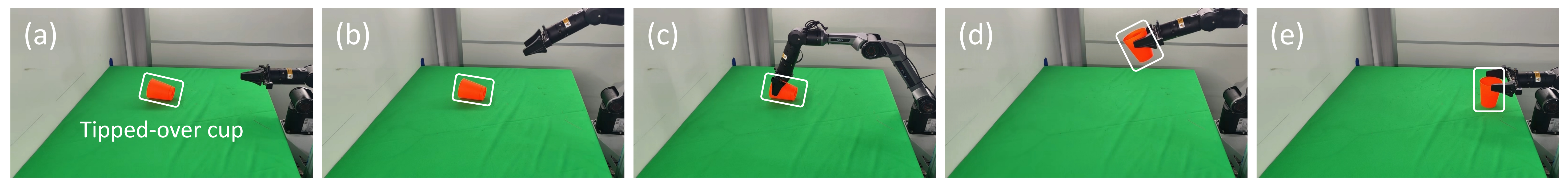} \\
  \vspace{1pt}
  
  \includegraphics[width=\linewidth]{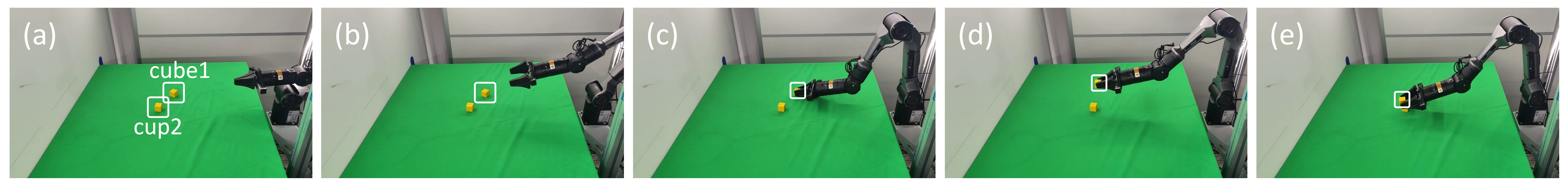}

  \caption{\textbf{Real-world Experiments.} The image sequence (top to bottom) illustrates the robot successfully performing three tasks: placing an object, uprighting a fallen cup, and stacking two blocks.}
  \label{fig:vis}
  \vspace{-2.0mm}
\end{figure}

\textbf{Results.}
We report the success rates across 15 real-world trials in Tab. \ref{tab:real_sr}. Consistent with our simulation findings, PocketDP3-base exhibits strong sim-to-real transferability. It outperforms the DP3 baseline by 15.3\% on average, validating that the efficiency gains from the DiM decoder do not come at the cost of generalization capability, even under the constraints of real-world noise.




\begin{table}[h]

    \centering
    \caption{\textbf{Real-world Experiments Success Rate.}}
    \label{tab:real_sr}
    \renewcommand{\arraystretch}{1.15}
    \setlength{\tabcolsep}{8pt}
    \begin{tabular}{lcc}
        \toprule
        Task &  DP3 & PocketDP3 \textbf{(Ours)} \\
        \midrule
        Place Object &  53.3 & \textbf{73.3} (\textbf{$\uparrow$20.0}) \\
        Adjust Bottle &  33.3 & \textbf{46.7} (\textbf{$\uparrow$13.4}) \\
        Stack Blocks Two &  6.7 & \textbf{20} (\textbf{$\uparrow$13.3}) \\
        \bottomrule
    \end{tabular}
    \vspace{-3.0mm}
\end{table}

\section{Discussion}
In this work, we introduced \textbf{PocketDP3}, a compact yet powerful 3D visuomotor policy. Building on the insight that, with a compact 3D representation, an oversized decoder may be redundant and may even introduce noise, we replace the parameter-heavy conditional U-Net decoder with a lightweight MLP-Mixer–based DiM block. Across a broad suite of simulation tasks, PocketDP3 achieves state-of-the-art performance with $<$1$\%$ of the parameters of prior methods and only two denoising steps, without requiring additional consistency-distillation training. Real-world experiments further validate its strong generalization, suggesting a practical path toward efficient 3D visuomotor control.

\subsection{Limitations}
\textbf{DiM Capacity Under Complex Conditioning.}
Our core conclusion builds on the premise that the encoder already provides a compact and information-rich scene representation. However, if the encoder outputs a higher-dimensional representation---for instance, with multi-view or multi-modal inputs---it may be necessary to increase the capacity of DiM to handle the more complex conditioning signal. 

\textbf{Fine-grained Planning Capability.}
Moreover, although the MLP-Mixer-style decoder offers extreme efficiency, it sacrifices the pixel-aligned spatial biases inherent in U-Nets. For tasks requiring fine-grained local geometric reasoning (e.g., insertion with sub-millimeter tolerances) or high-resolution spatial conditioning, the global mixing mechanism of PocketDP3 might face a "spatial bottleneck". Future work could explore hybrid architectures that reintroduce lightweight spatial attention for such precision-critical tasks.

\textbf{Underlying Mechanism of Two-Step Inference.}
Although we find that our model can perform inference with only two steps without sacrificing performance, the underlying reason remains unclear. We do not yet know whether this property arises primarily from the architecture or from the training procedure, nor whether it can be readily transferred to other related tasks (e.g., trajectory prediction) or methods.

\subsection{Future work}
\textbf{Efficient, Information-Rich Conditioning Encoders.}
One promising direction for future work is to design a more efficient encoder that maps scene conditions into a representation that is both sufficiently compact and information-rich. This includes stronger multi-modal fusion (e.g., vision, language, tactile) and temporal fusion over longer histories, which may ultimately be more critical to visuomotor policy performance than decoder scaling.

\textbf{Balancing Conditioning Capacity, Task Complexity, and Decoder Scale.}
In addition, it is worth systematically studying the trade-off among the information capacity of the conditioning representation, task complexity, and decoder capacity—i.e., how to balance these three factors to achieve an optimal overall design.

\newpage
\section*{Impact Statement}
This paper presents work whose goal is to advance the field of Machine
Learning. There are many potential societal consequences of our work, none
which we feel must be specifically highlighted here.

\bibliography{example_paper}
\bibliographystyle{icml2026}

\newpage
\appendix
\onecolumn
\section{Implementation Details} \label{app:impl}
In this section, we provide the essential hyperparameters required to reproduce our results. Detailed hyperparameter choices are listed in Tab. \ref{tab:hyperparameters}.
\begin{table}[h]
\centering
\caption{Hyperparameter Settings}
\label{tab:hyperparameters}
\setlength{\tabcolsep}{8pt}
\renewcommand{\arraystretch}{1.15}
\begin{tabular}{@{}llr@{}}
\toprule
\textbf{Category} & \textbf{Hyperparameter} & \textbf{Value} \\
\midrule
\multicolumn{3}{@{}l}{\textbf{Training}} \\
\midrule
& Batch size (Robotwin2.0) & 256 \\
& Batch size (Adroit \& MetaWorld) & 128 \\
& Num. epochs & 3000 \\
& Optimizer & AdamW \\
& Weight decay(Robotwin2.0) & $1\times10^{-6}$ \\
& Weight decay(Adroit \& MetaWorld)& $1\times10^{-8}$ \\
& LR scheduler & Cosine \\
& LR warmup steps & 500 \\
& Learning rate & $1\times10^{-4}$ \\
& Horizon (Robotwin2.0) & 8 \\
& Horizon (Adroit \& MetaWorld) & 16 \\
& Num. action steps (Robotwin2.0) & 6 \\
& Num. action steps (Adroit \& MetaWorld) & 8 \\
& Observation steps (Robotwin2.0) & 3 \\
& Observation steps (Adroit \& MetaWorld) & 2 \\
& Encoder output dim. & 64 \\
& Diffusion timestep dim. & 64 \\
& MLP expansion ratio. & 4 \\
\midrule
\multicolumn{3}{@{}l}{\textbf{Inference}} \\
\midrule
& Num. inference steps & 2 \\
& Num. train steps & 100 \\
\bottomrule
\end{tabular}
\end{table}


\end{document}